\documentclass[journal]{IEEEtran}

\usepackage{lineno,hyperref}
\usepackage{bm}
\usepackage{mathrsfs}
\usepackage{amsmath}
\usepackage{amssymb}
\usepackage{graphicx}
\usepackage{subfigure}
\usepackage{epstopdf}
\usepackage{xcolor}
\usepackage{wasysym}
\usepackage{footnote}
\usepackage{amssymb}
\usepackage{overpic}
\modulolinenumbers[5]

\ifCLASSINFOpdf

\else

\fi

\begin{document}

\title{Sparsity-Aware Robust Normalized Subband Adaptive Filtering algorithms based on Alternating Optimization}

\author{Yi~Yu,~\IEEEmembership{Member,~IEEE},
        ~Zongxin~Huang,
        ~Hongsen~He,~\IEEEmembership{Member,~IEEE},
        ~Yuriy Zakharov,~\IEEEmembership{Senior Member,~IEEE},
        ~and
        ~Rodrigo C. de Lamare,~\IEEEmembership{Senior Member,~IEEE}

\thanks{This work was supported in part by National Science Foundation of P.R. China (Nos. 61901400, 62071399), Doctoral Research Fund of Southwest University of Science and Technology in China (No. 19zx7122), Sichuan Science and Technology Program (No. 2021YFG0253), and Natural Science Foundation of Sichuan in 2022, China. The work of Y. Zakharov was supported in part by the U.K. EPSRC through Grants EP/R003297/1 and EP/V009591/1. Corresponding author: Hongsen He. }

\thanks{Y. Yu, Z. Huang and H. He are with School of Information Engineering, Robot Technology Used for Special Environment Key Laboratory of Sichuan Province, Southwest University of Science and Technology, Mianyang, 621010, China (e-mail: yuyi\_xyuan@163.com).}

\thanks{Y. Zakharov is with the Department of Electronics, University of York, York YO10 5DD, U.K. (e-mail: yury.zakharov@york.ac.uk).}

\thanks{R. C. de Lamare is with CETUC, PUC-Rio, Rio de Janeiro 22451-900, Brazil. (e-mail: delamare@cetuc.puc-rio.br).}
}


\maketitle

\begin{abstract}
This paper proposes a unified sparsity-aware robust normalized subband adaptive filtering (SA-RNSAF) algorithm for identification of sparse systems under impulsive noise. The proposed SA-RNSAF algorithm generalizes different algorithms by defining the robust criterion and sparsity-aware penalty. Furthermore, by alternating optimization of the parameters (AOP) of the algorithm, including the step-size and the sparsity penalty weight, we develop the AOP-SA-RNSAF algorithm, which not only exhibits fast convergence but also obtains low steady-state misadjustment for sparse systems. Simulations in various noise scenarios have verified that the proposed AOP-SA-RNSAF algorithm outperforms existing techniques.
\end{abstract}

\begin{IEEEkeywords}
Impulsive noises, subband adaptive filters, sparse systems, time-varying parameters.
\end{IEEEkeywords}

\IEEEpeerreviewmaketitle

\section{Introduction}

\IEEEPARstart{F}{or} highly correlated input signals, the normalized
subband adaptive filtering (NSAF)~\cite{lee2004improving} algorithm
provides faster convergence than the normalized least mean square
(NLMS) algorithm and retains comparable complexity. The NSAF
algorithm was proposed based on the multiband structure of subband
filters~\cite{lee2009subband}, which adjusts the fullband filter's
coefficients to remove the aliasing and band edge effects of the
conventional subband structure~\cite{lee2009subband}. However, in
practice the non-Gaussian noise with impulsive samples could
commonly happen such as in echo cancellation, underwater acoustics,
audio processing, and
communications~\cite{nikias1995signal,zimmermann2002analysis}, and
in this scenario, the NSAF performance degrades. To deal with
impulsive noises, several robust subband algorithms based on
different robust criteria were proposed,
see~\cite{kim2017delayless,huang2017combined,hur2016variable,zheng2020robust,yu2016two,yu2019m}
and references therein, and most of them can be unified as the NSAF
update with a specific scaling factor.

Furthermore, it is interesting to improve the adaptive filter
performance by exploiting the system sparsity. For example, the
impulse responses of propagation channels in underwater acoustic and
radio communications are usually
sparse~\cite{radecki2002echo,schreiber1995advanced}, only a few
coefficients of which are non-zero. Aiming at sparse systems,
existing examples are classified into the proportionate type and
sparsity-aware type. The family of proportionate NSAF (PNSAF)
algorithms~\cite{abadi2011family} assigns an individual gain to each
filter coefficient, which has faster convergence than the NSAF
algorithm. Later, robust PNSAF algorithms were also
presented~\cite{zheng2017robust,yu2019m} to deal with impulsive
noises. On the other hand, the family of sparsity-aware algorithms
incorporates the sparsity-aware penalty into the original NSAF's and
PNSAF's cost functions; as a result, sparsity-aware NSAF
(SA-NSAF)~\cite{yu2019sparsity,heydari2021improved} and
sparsity-aware PNSAF~\cite{puhan2019zero} algorithms were developed.
In sparse system identification, these sparsity-aware algorithms can
obtain better convergence and steady-state performance than their
original counterparts.

However, the superiority of sparsity-aware algorithms depends mainly
on the sparsity-penalty parameter, which is often chosen in an
exploratory way thus reducing the practicality of the algorithms.
Besides, they encounter the problem of choosing the step-size, which
controls the tradeoff between convergence rate and steady-state
misadjustment. Hence, adaptation techniques for the sparsity-penalty
and the step-size parameters are necessary. In the literature, they
are rarely discussed simultaneously regardless of the Gaussian noise
or impulsive noise scenarios. In~\cite{ji2020sparsity}, the variable
parameter SA-NSAF (VP-SA-NSAF) algorithm was proposed for the
Gaussian noise, in which these two parameters are jointly adapted
based on a model-driven method, but it requires knowledge of
variances of the subband noises. In~\cite{yu2021sparsity}, by
optimizing the parameters in the sparsity-aware
individual-weighting-factors-based sign subband adaptive filter
(S-IWF-SSAF) algorithm with the robustness in the impulsive noise,
the variable parameters S-IWF-SSAF (VP-S-IWF-SSAF) algorithm was
presented, while it lacks the generality in sparsity-aware subband
algorithms.

In this paper, we propose a unified sparsity-aware robust NSAF
(SA-RNSAF) framework to handle impulsive noises, which can result in
different algorithms by only changing the robustness criterion and
the sparsity-aware penalty. We then devise adaptive schemes for
adjusting the step-size and the sparsity-aware penalty weight, and
develop the alternating optimization of the parameters based
SA-RNSAF (AOP-SA-RNSAF) algorithm, with fast convergence and low
steady-state misadjustment for sparse systems.

\section{Statement of Problem and SA-RNSAF Algorithm}

\begin{figure}[htb]
    \centering
    \includegraphics[scale=0.42] {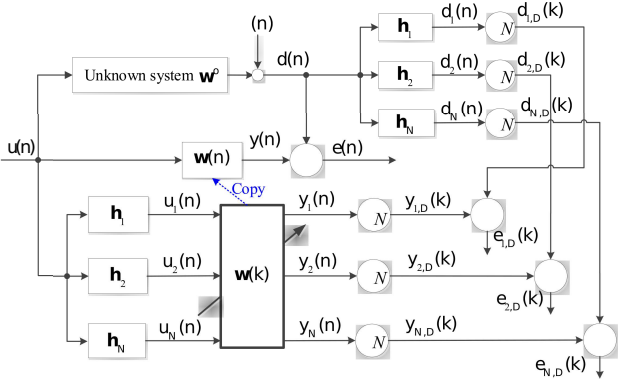}
    \vspace{-1em} \caption{Multiband structure of subband adaptive filter. }
    \label{Fig1}
\end{figure}

Consider a system identification problem. The relationship between
the input signal $u(n)$ and desired output signal $d(n)$ at time~$n$
is given by
\begin{align}
\label{001}
d(n) = \bm u^\text{T}(n) \bm w^o+\nu(n),
\end{align}
where the $M\times1$ vector $\bm w^o$ is the impulse response of the
sparse system that we want to identify, $\bm
u(n)=[u(n),u(n-1),...,u(n-M+1)]^\text{T}$ is the $M\times1$ input
vector, and $\nu(n)$ is the additive noise independent of $u(n)$.
For estimating~$\bm w^o$, the SAF with a coefficient vector~$\bm
w(k)$ is used, shown in Fig.~\ref{Fig1} with $N$ subbands, where $k$
denotes the sample index in the decimated domain. The input signal
$u(n)$ and the desired output signal~$d(n)$ are decomposed into
multiple subband signals $u_i(n)$ and $d_i(n)$ via the analysis
filters $\{\bm h_i\}_{i=1}^N$, respectively. For each subband input
signal~$u_i(n)$, the corresponding output of the fullband
filter~$\bm w(k)$ is $y_i(n)$. Then, both $d_i(n)$ and $y_i(n)$ are
critically decimated to yield signals $d_{i,D}(k)$ and $y_{i,D}(k)$,
respectively, with lower sampling rate, namely, $d_{i,D}(k)=d_i(kN)$
and $y_{i,D}(k)=\bm u_i^\text{T}(k)\bm w(k)$, where $\bm
u_i(k)=[u_i(kN),u(kN-1),...,u(kN-M+1)]^\text{T}$. By subtracting
$y_{i,D}(k)$ from $d_{i,D}(k)$, the decimated subband error signals
are obtained:
\begin{align}
e_{i,D}(k) = d_{i,D}(k)-\bm u_i^\text{T}(k)\bm w(k),\;i=1,2,...,N,
\label{002}
\end{align}
which are used to adjust the coefficient vector~$\bm
w(k)$~\footnote{In some applications, we could also eventually need
the output error $e(n)$ in the original time domain. To this end, we
obtain $\bm w(n)$ by copying $\bm w(k)$ for every $N$ input samples,
and then compute the output error by $e(n) = d(n)-\bm u^\text{T}(n)
\bm w(n)$. }.

In practice, the additive noise~$\nu(n)$ can be non-Gaussian
consisting of Gaussian and impulsive components
~\cite{jidf,spa,intadap,mbdf,jio,jiols,jiomimo,sjidf,ccmmwf,tds,mfdf,l1stap,mberdf,jio_lcmv,locsme,smtvb,ccmrls,dce,itic,jiostap,aifir,ccmmimo,vsscmv,bfidd,mbsic,wlmwf,bbprec,okspme,rdrcb,smce,armo,wljio,saap,vfap,saalt,mcg,sintprec,stmfdf,1bitidd,jpais,did,rrmber,memd,jiodf,baplnc,als,vssccm,doaalrd,jidfecho,dcg,rccm,ccmavf,mberrr,damdc,smjio,saabf,arh,lsomp,jrpaalt,smccm,vssccm2,vffccm,sor,aaidd,lrcc,kaesprit,lcdcd,smbeam,ccmjio,wlccm,dlmme,listmtc,smcg},
. Hence, for the identification of a sparse vector~$\bm w^o$ in the
presence of impulsive noise, we define the following minimization
problem:
\begin{equation}
\label{003}
\begin{array}{rcl}
\begin{aligned}
\arg \min_{\bm w(k+1)} &\left[ ||\bm w(k+1) - \bm w(k)||_2^2 +\rho f(\bm w(k+1)) \right],\\
\end{aligned}
\end{array}
\end{equation}
subject to
\begin{subequations} \label{eq:4}
    \begin{align}
e_{p,i}(k) =& \left[ 1 - \mu \phi_i(k) \right] e_{i,D}(k),
    \label{eq:4a}\\
    \phi_i(k)=&\frac{\varphi'(e_{i,D}(k))}{e_{i,D}(k)}, \label{eq:4b}
    \end{align}
\end{subequations}
for subbands $i=1,...,N$, where $e_{p,i}(k)=d_{i,D}(k)-\bm u_i^\text{T}(k)\bm w(k+1)$ denotes the \textit{a posteriori} decimated subband error, $\mu>0$ will be called the step-size in the sequel, and $\phi_i(k)$ is called the scaling factor of the $i$-th subband. In~\eqref{003}, $f(\bm w)$ is a sparsity-aware penalty function and $\rho>0$ is the weight of this penalty term. In~\eqref{eq:4b}, $\varphi'(e) \triangleq \frac{\partial \varphi(e)}{\partial e}$, where $\varphi(e)\geq 0$ is an even function of variable~$e$, defining the robustness to impulsive noise.

By using the Lagrange multiplier method, we obtain the solution of~\eqref{003} subject to~\eqref{eq:4a} as
\begin{equation}
\label{005}
\begin{array}{rcl}
\begin{aligned}
\bm w&(k+1) =\bm w(k) + \mu \sum_{i=1}^{N} \phi_i(k) \frac{e_{i,D}(k)\bm u_i(k)}{||\bm u_i(k)||_2^2} - \\
&\rho \left[ f'(\bm w(k+1)) - \sum_{i=1}^{N} \frac{\bm u_i(k) \bm u_i^\text{T}(k)}{||\bm u_i(k)||_2^2} f'(\bm w(k+1)) \right].
\end{aligned}
\end{array}
\end{equation}
Note that the derivation of~\eqref{005} also uses an approximation
in the SAF, that is~$\bm u_i^\text{T}(k) \bm u_j(k)\approx 0$ for
$i\neq j$~\cite{lee2004improving}. Then, by introducing an
intermediate estimate~$\bm \psi(k)$, we propose to
implement~\eqref{005} in two steps:
\begin{subequations} \label{eq:6}
    \begin{align}
    \bm \psi(k) = &\bm w(k) + \mu \sum_{i=1}^{N} \phi_i(k) \frac{e_{i,D}(k)\bm u_i(k)}{||\bm u_i(k)||_2^2},
    \label{eq:6a}\\
    \bm w(k+1) =& \bm \psi(k) - \rho \bm P(k), \label{eq:6b}
    \end{align}
\end{subequations}
where
\begin{equation}
\label{007}
\begin{array}{rcl}
\begin{aligned}
\bm P(k) = f'(\bm \psi(k)) - \sum_{i=1}^{N} \frac{\bm u_i(k) \bm u_i^\text{T}(k)}{||\bm u_i(k)||_2^2} f'(\bm \psi(k)).
\end{aligned}
\end{array}
\end{equation}
This completes the derivation of the update for the SA-RNSAF
algorithm. In this algorithm, the steps~\eqref{eq:6a}
and~\eqref{eq:6b} have their own roles. The former behaves like the
RNSAF algorithm to obtain a coarse estimate~$\bm \psi(k)$ of~the
sparse vector $\bm w^o$ in impulsive noise. Subsequently, the
step~\eqref{eq:6b} forces the inactive coefficients in $\bm \psi(k)$
to zero, thus obtaining a more accurate sparse estimate~$\bm
w(k+1)$.

It is noteworthy that the parameters~$\mu$ and $\rho$ control the
SA-RNSAF's performance. Specifically, the step-size~$\mu$ controls
the convergence rate and steady-state misadjustment of the
algorithm. Moreover, the SA-RNSAF algorithm can be superior to the
RNSAF algorithm when dealing with sparse systems, but $\rho$ must be
chosen within a theoretically existing range while this range is
unpredictable actually (see Remark~1 below). As such, we will derive
adaptive recursions for adjusting $\mu$ and $\rho$. However, it is
challenging to solve the global optimization problem on $\mu$ and
$\rho$, as~\eqref{eq:6a} and \eqref{eq:6b} depend on each other.
Interestingly, $\mu$ and $\rho$ mainly affect the
steps~\eqref{eq:6a} and \eqref{eq:6b}, respectively, thus we can use
alternating optimization~\cite{hong2016convergence} to solve this
global optimization problem. Accordingly, the adaptations of $\mu$
and $\rho$ will be designed independently according to~\eqref{eq:6a}
and~\eqref{eq:6b}, respectively.

\section{Proposed AOP-SA-RNSAF Algorithm}

By using the band-dependent variable step-size (VSS) $\mu_i(k)$ and
$\rho(k)$ instead of some fixed values, we rearrange~\eqref{eq:6a}
and \eqref{eq:6b} as follows:
\begin{subequations} \label{eq:8}
    \begin{align}
    \bm \psi(k) = &\bm w(k) + \sum_{i=1}^{N} \mu_i(k) \phi_i(k) \frac{e_{i,D}(k)\bm u_i(k)}{||\bm u_i(k)||_2^2},
    \label{eq:8a}\\
    \bm w(k+1) =& \bm \psi(k) - \rho(k) \bm P(k). \label{eq:8b}
    \end{align}
\end{subequations}
\subsection{Adaptation of the step-size}
By subtracting~\eqref{eq:8a} from $\bm w^o$, we obtain
\begin{equation}
\label{009}
\begin{array}{rcl}
\begin{aligned}
\widetilde{\bm \psi}(k) = &\widetilde{\bm w}(k) - \sum_{i=1}^{N} \mu_i(k) \phi_i(k) \frac{e_{i,D}(k)\bm u_i(k)}{||\bm u_i(k)||_2^2},
\end{aligned}
\end{array}
\end{equation}
where $\widetilde{\bm w}(k)=\bm w^o - \bm w(k)$ and $\widetilde{\bm \psi}(k)=\bm w^o - \bm \psi(k)$ define the deviation vectors for the final estimate~$\bm w(k)$ and the intermediate estimate $\bm \psi(k)$ with respect to the true value. By pre-multiplying~$\bm u_i^\text{T}(k)$ on both sides of~\eqref{009} and applying the approximation~$\bm u_i^\text{T}(k) \bm u_j(k)\approx 0$ for $i\neq j$ again, it is established that
\begin{equation}
\label{010}
\begin{array}{rcl}
\begin{aligned}
e_{\varepsilon,i}(k) = \left[ 1 -\mu_i(k) \phi_i(k) \right]  e_{i,D}(k)
\end{aligned}
\end{array}
\end{equation}
for $i=1,2,...,N$, where $e_{\varepsilon,i}(k) = d_{i,D}(k)-\bm u_i^\text{T}(k) \bm \psi(k)$ defines the intermediate \textit{a posteriori} error at the $i$-th subband resulting from the step~\eqref{eq:6a}. By squaring both sides of~\eqref{010} and taking the expectations over all the terms, the following relation is obtained:
\begin{equation}
\label{011}
\begin{array}{rcl}
\begin{aligned}
\text{E}\{e_{\varepsilon,i}^2(k)\} = \left[ 1 -\mu_i(k) \phi_i(k) \right]^2  \text{E}\{e_{i,D}^2(k)\},
\end{aligned}
\end{array}
\end{equation}
where $\text{E}\{\cdot\}$ denotes the mathematical expectation. In~\eqref{011}, a common assumption is used that the step-size $\mu_i(k)$ and the scaling factors $\{\phi_i(k)\}_{i=1}^N$ are deterministic at iteration~$k$ in contrast with the randomness of $e_{i,D}(k)$~\cite{ni2010variable,hur2016variable}.

Motivated by~\cite{ni2010variable}, we wish to compute the subband step-sizes in such a way that $\text{E}\{e_{\varepsilon,i}^2(k)\}=\sigma_{\nu,i}^2$, $i=1,2,...,N$, which means that the powers of the intermediate \textit{a posteriori} subband errors always equal those of the subband noises, where $\sigma_{\nu,i}^2\triangleq\text{E}\{\nu_{i,D}^2(k)\}$ denotes the power of the $i$-th subband noise excluding impulsive interferences. on this requirement, then from~\eqref{011} we can obtain the following equation:
\begin{equation}
\label{012}
\begin{array}{rcl}
\begin{aligned}
\mu_i(k) \phi_i(k) = 1- \sqrt{\frac{\sigma_{\nu,i}^2}{\sigma_{e_{i,D}}^2(k)}},
\end{aligned}
\end{array}
\end{equation}
where $\sigma_{e_{i,D}}^2(k)\triangleq \text{E}\{e_{i,D}^2(k)\}$ indicates the power of $e_{i,D}(k)$ without impulsive noises. For robust adaptive algorithms with the scaling factors, there is a common property~\cite{huang2017combined,hur2016variable} that when impulsive noises happen, the scaling factors $\phi_i(k)$ will become very small (close to zero), thereby preventing the adaptation~\eqref{eq:8a} from the interference caused by impulsive noises. If the impulsive noise is absent, $\phi_i(k)$ will approximately equal one to ensure fast convergence. As such, we can change~\eqref{012} to
\begin{equation}
\label{013}
\begin{array}{rcl}
\begin{aligned}
\mu_i(k) = 1- \sqrt{\frac{\sigma_{\nu,i}^2}{\sigma_{e_{i,D}}^2(k)}}.
\end{aligned}
\end{array}
\end{equation}
To implement \eqref{013}, the statistical quantities $\sigma_{e_{i,D}}^2(k)$ and $\sigma_{\nu,i}^2$ are replaced with their estimates $\hat{\sigma}_{e_{i,D}}^2(k)$ and $\hat{\sigma}_{\nu,i}^2(k)$, respectively. Specifically, $\hat{\sigma}_{e_{i,D}}^2(k)$ is calculated in an exponential window way as
\begin{equation}
\label{014}
\begin{array}{rcl}
\begin{aligned}
\hat{\sigma}_{e_{i,D}}^2(k) = \zeta \hat{\sigma}_{e_{i,D}}^2(k-1) + (1-\zeta)\phi_i^2(k) e_{i,D}^2(k),
\end{aligned}
\end{array}
\end{equation}
where $\zeta$ is a weighting factor often chosen as $\zeta=1-1/(\kappa M)$ with $\kappa \geq1$. Similar to~\cite{ni2010variable}, $\hat{\sigma}_{\nu,i}^2(k)$ is calculated by the following equations:
\begin{subequations} \label{eq:15}
    \begin{align}
    \hat{\sigma}_{u_{i}}^2(k) &= \zeta \hat{\sigma}_{u_{i}}^2(k-1) + (1-\zeta) u_{i}^2(kN),
    \label{eq:15a}\\
    \hat{\bm r}_{ue_{i}}(k) &= \zeta \hat{\bm r}_{ue_{i}}(k-1) + (1-\zeta) \phi_i(k) e_{i,D}(k) \bm u_{i}(k), \label{eq:15b}\\
    \hat{\sigma}_{\nu,i}^2(k) & = \hat{\sigma}_{e_{i,D}}^2(k) - \frac{||\hat{\bm r}_{ue_{i}}(k)||_2^2}{\hat{\sigma}_{u_{i}}^2(k) + \epsilon_1}, \label{eq:15c}
    \end{align}
\end{subequations}
where $\epsilon_1$ is a small positive number (e.g., $10^{-5}$). Note that, we introduce the scaling factor $\phi_i(k)$ in (14) and (15b) for each subband to suppress impulsive noises.

Accordingly, \eqref{013} can be rewritten as
\begin{equation}
\label{016}
\begin{array}{rcl}
\begin{aligned}
\mu_i(k) = 1- \sqrt{\frac{\hat{\sigma}_{\nu,i}^2(k)}{\hat{\sigma}_{e_{i,D}}^2(k) + \epsilon_2}},
\end{aligned}
\end{array}
\end{equation}
where $\epsilon_2$ is also a small positive number. It is stressed that the estimated values of multiple statistical quantities are used in~\eqref{eq:15c}, and thus $\hat{\sigma}_{\nu,i}^2(k)$ could be negative at some iterations. To avoid this, we add the step $\hat{\sigma}_{\nu,i}^2(k)\leftarrow \hat{\sigma}_{\nu,i}^2(k-1)$ after~\eqref{eq:15c}.
\subsection{Adaptation of the sparsity penalty weight}
By subtracting~\eqref{eq:8b} from $\bm w^o$, we obtain
\begin{equation}
\label{017}
\begin{array}{rcl}
\begin{aligned}
\widetilde{\bm w}(k+1) = \widetilde{\bm \psi}(k) + \rho(k) \bm P(k).
\end{aligned}
\end{array}
\end{equation}
By pre-multiplying both sides of~\eqref{017} by their transpose, we obtain
\begin{equation}
\label{018}
\begin{array}{rcl}
\begin{aligned}
||\widetilde{\bm w}&(k+1)||_2^2 = ||\widetilde{\bm \psi}(k)||_2^2 + \triangle(k),
\end{aligned}
\end{array}
\end{equation}
where
 \begin{equation}
 \label{019}
 \begin{array}{rcl}
 \begin{aligned}
 \triangle(k) = 2 \rho(k) \widetilde{\bm \psi}^\text{T}(k)\bm P(k) + \rho^2(k) ||\bm P(k)||_2^2.
 \end{aligned}
 \end{array}
 \end{equation}

Remark~1: \eqref{018} clearly reveals that the proposed SA-RNSAF algorithm will outperform the RNSAF algorithm for identifying sparse systems, if and only if $\triangle(k)<0$ holds\footnote{Following a derivation similar to that in~Appendix~D in \cite{yu2021sparsity}, $\triangle(k)<0$ is likely to be true as long as $\bm w^o$ is sparse.}. It follows that $\rho(k)$ should satisfy the inequality
  \begin{equation}
 \label{020}
 \begin{array}{rcl}
 \begin{aligned}
0<\rho(k) < 2 \frac{ [\bm \psi(k) - \bm w^o]^\text{T} \bm P(k)}{||\bm P(k)||_2^2}.
 \end{aligned}
 \end{array}
 \end{equation}
Moreover, since $\triangle(k)$ is the quadratic function of~$\rho(k)$, there is an optimal $\rho(k)$ such that $\triangle(k)$ arrives at the negative maximum value. Consequently, the optimal $\rho(k)$ is given as
\begin{equation}
\label{021}
\begin{array}{rcl}
\begin{aligned}
\rho_\text{opt}(k) = \frac{ [\bm \psi(k) - \bm w^o]^\text{T} \bm P(k)}{||\bm P(k)||_2^2}.
\end{aligned}
\end{array}
\end{equation}

Although Remark~1 states that the relations~\eqref{020} and~\eqref{021} are existing in sparse systems, they are incalculable due to the fact that the sparse vector~$\bm w^o$ is unknown. To solve this problem, we use the previous estimate $\bm w(k)$ to approximate $\bm w^o$, then~\eqref{021} can be reformulated as
\begin{equation}
\label{022}
\begin{array}{rcl}
\begin{aligned}
\hat{\rho}_\text{opt}(k) = \max \left\lbrace \frac{ [\bm \psi(k) - \bm w(k)]^\text{T} \bm P(k)}{||\bm P(k)||_2^2},0\right\rbrace ,
\end{aligned}
\end{array}
\end{equation}
where $\hat{\rho}_\text{opt}(k)$ is set to zero at $k=0$.

The recursion~\eqref{eq:8} equipped with $\mu_i(k)$ in~\eqref{016} and $\hat{\rho}_\text{opt}(k)$ in~\eqref{022} constitutes the proposed AOP-SA-RNSAF algorithm.

Remark~2: The proposed AOP-SA-RNSAF update generalizes different algorithms, depending on the choice of $\varphi(e)$ in~\eqref{eq:4b} and $f(\bm w)$ in~\eqref{003}. In the literature, several robust criteria against impulsive noises~\cite{zheng2017robust,yu2016two,huang2017combined,hur2016variable,yu2019m} defined by $\varphi(e)$ and sparsity-aware penalties~\cite{yu2021sparsity,heydari2021improved,ji2020sparsity,yu2019sparsity,de2014sparsity} defined by $f(\bm w)$ have been studied, which can be applied in the AOP-SA-RNSAF. Nevertheless, this paper does not consider the effect of different choices of $\varphi(e)$ and/or $f(\bm w)$, which is worth studying in future work. Note that, when setting $\varphi(e)=\frac{1}{2}e^2$, the proposed algorithm is called the alternating optimization of parameters based SA-NSAF (AOP-SA-NSAF) suited for Gaussian noise environments, which is a sparsity-aware variant of the VSS-NSAF algorithm presented in~\cite{ni2010variable}.

Remark~3: By firstly computing the inner product~$\bm u_i^\text{T}(k)f'(\bm \psi(k))$ in~$\bm P(k)$, and then calculating~$\bm P(k)$ only requires $2M$ multiplications, $2M-M/N$ additions, and 1 division. Therefore, the complexity of the proposed AOP-SRNSAF algorithm is still low with $\mathcal{O}(M)$ arithmetic operations per input sample.
\section{Simulation Results}
To evaluate the proposed AOP-SA-RNSAF algorithm, simulations are conducted to identify the acoustic echo paths with $M=512$ taps. The sparsenesses, defined as $\chi(\bm w^o) = \frac{M}{M-\sqrt{M}} \left(1-\frac{||w^o||_1}{\sqrt{M}||w^o||_2} \right)$, of two echo paths are $\chi(\bm w^o_1)=0.9357$  (sparse)~\cite{yu2021sparsity} and $\chi(\bm w^o_2)=0.3663$ (dispersive or non-sparse)~\cite{zheng2017robust}, respectively. The length of the adaptive filters is the same as that of $\bm w^o$. The correlated input signal $u(n)$ is a first-order autoregressive (AR) process with the pole at 0.9, generated by filtering a white Gaussian noise with zero-mean and unit variance. The analysis filters~$\{\bm h_i\}_{i=0}^{N-1}$ for decomposing signals $d(n)$ and $u(n)$ are obtained by cosine-modulated filter banks, where the length of the prototype filter for $N=4$ subbands is 33 to obtain 60 dB stopband attenuation. The high stopband attenuation is to guarantee that adjacent analysis filters have almost no overlap and the cross-correlation of nonadjacent subbands is negligible~\cite{lee2009subband}. The normalized mean square deviation (NMSD), defined as $\text{E}\{||\bm w(n)-\bm w^o||_2^2/||\bm w^o||_2^2\}$, is the performance measure. All the results are the average over 50 independent runs.

For the AOP-SA-RNSAF algorithm, we use the modified Huber~(MH) function for $\varphi(e)$ and the log-penalty for $f(\bm \psi(k))$. The MH function is formulated as $\varphi(e_{i,D}(k))=e_{i,D}^2(k)/2$ if $|e_{i,D}(k)| < \xi_i$ and $\varphi(e_{i,D}(k))=0$  if $|e_{i,D}(k)| \geq \xi_i$~\cite{yu2019m}, where $\xi_i$ is the threshold. Accordingly, when $|e_{i,D}(k)| \geq \xi_i$ (usually impulsive noises occur), then the scaling factor in~\eqref{eq:4b} is $\phi_i(k)=0$, which makes the adaptation step~\eqref{eq:8a} freeze to suppress impulsive interferences; otherwise, $\phi_i(k)=1$, which retains fast convergence. Note that, the threshold $\xi_i$ for each subband $i$ is set to $\xi_i = 2.576 \hat{\sigma}_{\varepsilon,i}(k)$, where $\hat{\sigma}_{\varepsilon,i}^2(k)$ is the variance of $e_{i,D}(k)$ excluding impulsive samples. $\hat{\sigma}_{\varepsilon,i}^2(k)$ is computed by~$\hat{\sigma}_{\varepsilon,i}^2(k) = \lambda \hat{\sigma}_{\varepsilon,i}^2(k-1) + c_\sigma(1-\lambda) \text{med}(\bm a_{\varepsilon,i})$, where $\lambda\in (0.9, 1)$ is the forgetting factor (but $\lambda=0$ at $k=0$), $\text{med}(\cdot)$ denotes the median operator to remove outliers in the data window $\bm a_{\varepsilon,i}=[e_{i,D}^2(k),e_{i,D}^2(k-1),...,e_{i,D}^2(k-N_w+1)]$ with a length of $N_w$, and $c_\sigma=1.483(1+5/(N_w-1))$ is the correction factor. The log-penalty is given as~$f(\bm \psi_k) = \sum_{m=1}^{M} \ln(1+|\psi_{m,k}|/\theta)$~\cite{yu2021sparsity} which characterizes the sparsity of systems, where $\psi_{m,k}$ is the $m$-th element of $\bm \psi_k$, and the shrinkage factor~$\theta>0$ cuts apart inactive and active entries in~$\bm \psi_k$. Thus, $f'(\bm \psi_{m,k})$ in~\eqref{007} is computed element-wise as~$f'(\psi_{m,k}) = \frac{\text{sgn}(\psi_{m,k})}{\theta + |\psi_{m,k}|}, m=1,...,M$. In our simulations, the additive noise $\nu(n)$ is described by the symmetric $\alpha$-stable process, also called the $\alpha$-stable noise, whose characteristic function is formulated as $\phi(t)=\exp(-\gamma \lvert t \lvert^\alpha)$~\cite{nikias1995signal}. The parameter~$\alpha \in (0,2]$ represents the impulsiveness of the noise that for smaller $\alpha$ leads to stronger impulsive noises, and $\gamma>0$ behaves like the variance of the Gaussian density. In particular, it reduces to the Gaussian noise for the case of~$\alpha=2$. In the following simulations, we set $\gamma=0.02$.

\begin{figure}[htb]
    \centering
    \includegraphics[scale=0.5] {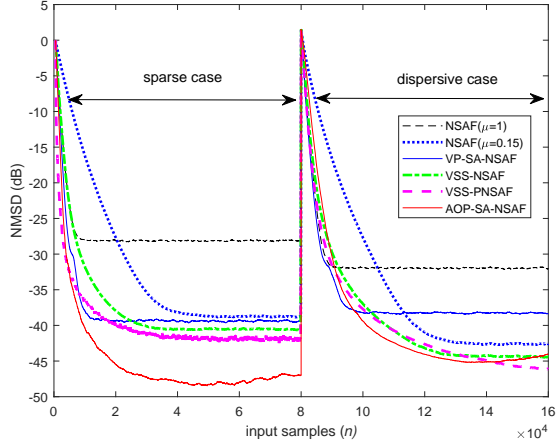}
    \vspace{-1em} \caption{NMSD performance of NSAF-type algorithms in the Gaussian noise. The parameters of algorithms are listed as follows: $\eta=0.99$, $\lambda=0.95$, and $\mu_{\max}=1$ for VP-SA-NSAF; $\tau=3$ for VSS-NSAF; $\tau=5$ and $\zeta=0$ for VSS-PNSAF; $\kappa=6$ for AOP-SA-NSAF. }
    \label{Fig3}
 \end{figure}

Example~1: the impulsive noise is absent, i.e., $\alpha=2$. The proposed AOP-SA-NSAF algorithm in Remark~2 is compared with the NSAF, VP-SA-NSAF~\cite{ji2020sparsity}, VSS-NSAF, and VSS-PNSAF algorithms in Fig.~\ref{Fig3}, where both VSS-NSAF and VSS-PNSAF are obtained from~\cite{yu2019m} but in the Gaussian noise we reset~$\varphi(e)=\frac{1}{2}e^2$ instead of using the MH function. For a fair evaluation, we select the log-penalty parameter $\theta=0.005$ for all the sparsity-aware algorithms. As can be seen, the VSS-NSAF algorithm obtains fast convergence and low steady-state misadjustment, which overcomes the trade-off problem in the NSAF algorithm. By considering the sparsity of the underlying system, both VP-SA-NSAF and VSS-PNSAF algorithms further improve the convergence rate. As compared to the VSS-PNSAF algorithm, the proposed AOP-SNSAF algorithm shows slower initial convergence, but it achieves higher reduction in the steady-state misadjustment.
\vspace{-0.5cm}

\begin{figure}[htb]
    \centering
    \includegraphics[scale=0.5] {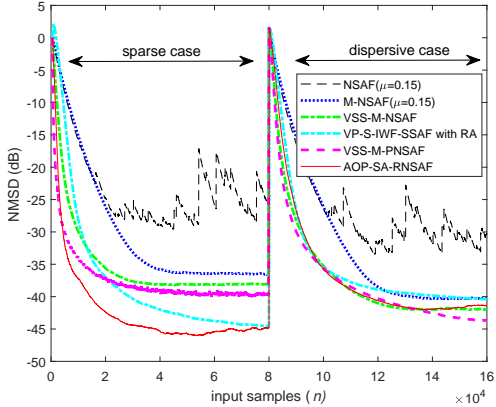}
    \hspace{2cm}\caption{NMSD performance of NSAF-type algorithms in the $\alpha$-stable noise. The parameters of algorithms are listed as follows: $\mu_{\min}=10^{-5}$, $\tau=2$, and $\chi=1$ for VP-S-IWF-SSAF; $\tau=3$ for VSS-M-NSAF; $\tau=5$ and $\zeta=0$ for VSS-M-PNSAF; $\kappa=6$ for AOP-SA-RNSAF.}
    \label{Fig4}
\end{figure}

Example~2: $\alpha=1.6$ displays the presence of impulsive noises. Fig.~\ref{Fig4} depicts the NMSD performance of the NSAF, M-NSAF~\cite{yu2019m}, VSS-M-NSAF~\cite{yu2019m}, VSS-M-PNSAF~\cite{yu2019m}, VP-IWF-SSAF with RA~\cite{yu2021sparsity}, and the proposed AOP-SA-RNSAF algorithms\footnote{Since the variance of the $\alpha$-stable noise is nonexistent, here we do not show the performance of the VP-SA-NSAF algorithm.}. For the M-estimate based algorithms, we choose the common M-estimate parameters $\lambda=0.99$ and $N_w =20$. It is seen that the NSAF algorithm shows poor misadjustment in the $\alpha$-stable noise, and other algorithms exhibit robust convergence. Among these robust algorithms, the proposed AOP-SA-RNSAF algorithm is the best choice for identifying sparse systems, due to the fact that it has lower steady-state misadjustment than the VSS-M-PNSAF and VP-S-IWF-SSAF with RA algorithms, even if it has a slightly slower initial convergence than the VSS-M-PNSAF algorithm.

It can be seen in Figs.~\ref{Fig3} and~\ref{Fig4} that, after $\bm w^o$ becomes dispersive at the middle of input samples, the proportionate-type (i.e., VSS-PNSAF, VSS-PNSAF) and sparsity-aware type (i.e., VP-SA-NSAF, AOP-SA-NSAF, AOP-SA-RNSAF) algorithms still show almost the same performance as the competing algorithms (i.e., VSS-NSAF and VSS-M-NSAF) in both Gaussian and $\alpha$-stable noise scenarios. In addition, as $\alpha$ decreases from 2 to 1.6, the steady-state misadjustment of the proposed AOP-SA-RNSAF algorithm increases, but this algorithm is still convergent.
\section{Conclusion}
In this paper, a unified SA-RNSAF framework for algorithms was developed for identifying sparse systems in impulsive noise environments. By replacing directly the specified robustness criterion and sparsity-aware penalty, it can yield different SA-RNSAF algorithms. We then developed adaptive techniques for the step-size and the sparsity penalty weight in the SA-RNSAF algorithm, thus arriving at the AOP-SA-RNSAF algorithm with a further performance improvement in terms of the convergence rate and steady-state misadjustment. Simulations for the sparse system identification have demonstrated the effectiveness of the proposed algorithms.

\ifCLASSOPTIONcaptionsoff
  \newpage
\fi

\newpage
\bibliographystyle{IEEEtran}
\bibliography{IEEEabrv,mybibfile}





%








\end{document}